\documentclass[11pt]{article}
\usepackage{times}
\usepackage{latexsym}
\usepackage[T1]{fontenc}
\usepackage[utf8]{inputenc}
\usepackage{microtype}
\usepackage{courier}
\usepackage{graphicx}
\usepackage{amsmath,amssymb}
\usepackage{booktabs}
\usepackage[normalem]{ulem}
\usepackage{tabularx}
\usepackage[preprint]{acl}

\title{REA: From Correct LLM Judgments to Safe Decision Delivery}

\author{
Jingxiang Meng\textsuperscript{1}\thanks{Equal contribution.},
Aofan Liu\textsuperscript{2}\footnotemark[1],
Yaoshuai Ma\textsuperscript{3}
\\
Fangxin Liu\textsuperscript{4}\thanks{Corresponding authors.},
Yongbiao Chen\textsuperscript{4}\footnotemark[2]
\\
\textsuperscript{1}University of Chicago\\
\textsuperscript{2}Peking University\\
\textsuperscript{3}Southern University of Science and Technology\\
\textsuperscript{4}Shanghai Jiao Tong University\\
\texttt{liufangxin@sjtu.edu.cn},
\texttt{chenyongbiao0319@sjtu.edu.cn}
}
\begin{document}

\maketitle

\begin{abstract}
Generative safety guards must classify requests correctly and return outputs that applications can parse. Reasoning guards can improve classification accuracy while producing outputs that applications cannot parse. When applications cannot extract a correct prediction, fallback handling can allow harmful requests or block safe ones. We introduce \textbf{Resolve--Express--Act (REA)}, a framework that traces judgments through parsing and fallback to application actions. Experiments on classifiers and safety guards show that correct predictions can appear in incorrectly formatted outputs. Parsing the safety verdict preserves the reasoning guard's accuracy advantage, while stricter explanation requirements reject additional correct predictions. Accepting formatting variations recovers most rejected correct predictions. Held-out tests confirm that output requirements and generation limits affect whether correct predictions can be parsed. In our fail-open replay, matching the parser to the output and fixing formatting reduced the fraction of harmful requests allowed despite correct guard judgments from 95.5\% to 0\%. A fixed-output comparison shows that model-based extraction can preserve guard verdicts, but is not automatically reliable; deterministic normalization already repairs the observed native-format failures.
Fallback replay shows that the same correct prediction can lead to different safety actions depending on the parser and fallback policy, including allowing requests the guard correctly identifies as harmful. 
\end{abstract}

\section{Introduction}

Generative safety guards classify requests and responses to support Allow or Block actions. Llama Guard and WildGuard express verdicts as text with prescribed vocabularies and formats~\citep{inan2023llamaguard,han2024wildguard}. {As guards add reasoning to their verdicts, applications must handle more complex outputs. }The application must extract the verdict and handle unparseable outputs. Reliability therefore depends on classification, extraction, and fallback.

A guard may correctly identify a harmful request but place its verdict after an explanation. A parser expecting a bare first-line label rejects it: fail-open then allows the request, while fail-closed blocks. This mismatch can arise when a reasoning guard replaces a label-style guard without updating the parser, or when an instruction rewrite drops the output-form clause.

GuardReasoner generates reasoning before its verdict and reports improved classification~\citep{liu2025guardreasoner}; rationale-based methods also improve classification robustness~\citep{chen2022rationalization,kavumba2023explanation,he2024xicl}. These outputs introduce structure for the consumer to handle. A verdict-only consumer reads the final answer block, whereas a trace consumer also reads task boundaries, step headings, and bodies. The same{ semantically correct} response can satisfy one contract and fail another.

Existing evaluation{s} {}assess {}{classification and instruction following}. Safety benchmarks measure harmfulness detection, refusal behavior, and robustness to adversarial requests~\citep{bassani2024guardbench,mazeika2024harmbench}. Instruction-following evaluations test whether outputs respect specified constraints~\citep{zhou2023ifeval}, while prompt-sensitivity studies show that extraction rules can change the scores assigned to semantically correct responses~\citep{hua2025flaw}. {}{Although t}hese {}{evaluations assess judgment quality and output compliance,} a practical question {remains }for deployment: how much of a guard's classification advantage survives the output contract, and what action follows when that contract fails? Answering it requires following the same response from its expressed judgment through extraction and fallback.

{}{To answer this question, w}e model this path as Resolve--Express--Act: identify the expressed judgment, test its delivery under the consumer contract, and apply execution and fallback rules. Semantic accuracy $A$ and conditional contract compliance $\kappa$ separate wrong judgments from correct judgments lost at the interface. Their product $A\kappa$ measures correct and executable decisions, quantifying how much accuracy the consumer retains.

Our classification experiments isolate delivery behavior by comparing instruction renderings over fixed content and label spaces. All semantic judgments are correct, yet 19.4\% of the responses violate the bare-label contract. The failures predominantly expand into sentences or explanations, and they concentrate under instructions without an explicit output-form clause. Applying different extractors to these responses reveals which output forms each consumer can recover. On XSTest safety requests, we then compare a label-style guard with a reasoning guard and evaluate their output contracts. GuardReasoner has higher classification accuracy, and its strict native-answer consumer delivers 93.0\% correct and executable decisions. Requiring a structured reasoning trace reduces this fraction to 80.1\%; accepting trace-format variants recovers most of the difference. The {}{output contract therefore determines how much of} the guard{'s accuracy advantage} {}{the application retains}.

\begin{figure*}[t]
\centering
\includegraphics[width=\textwidth]{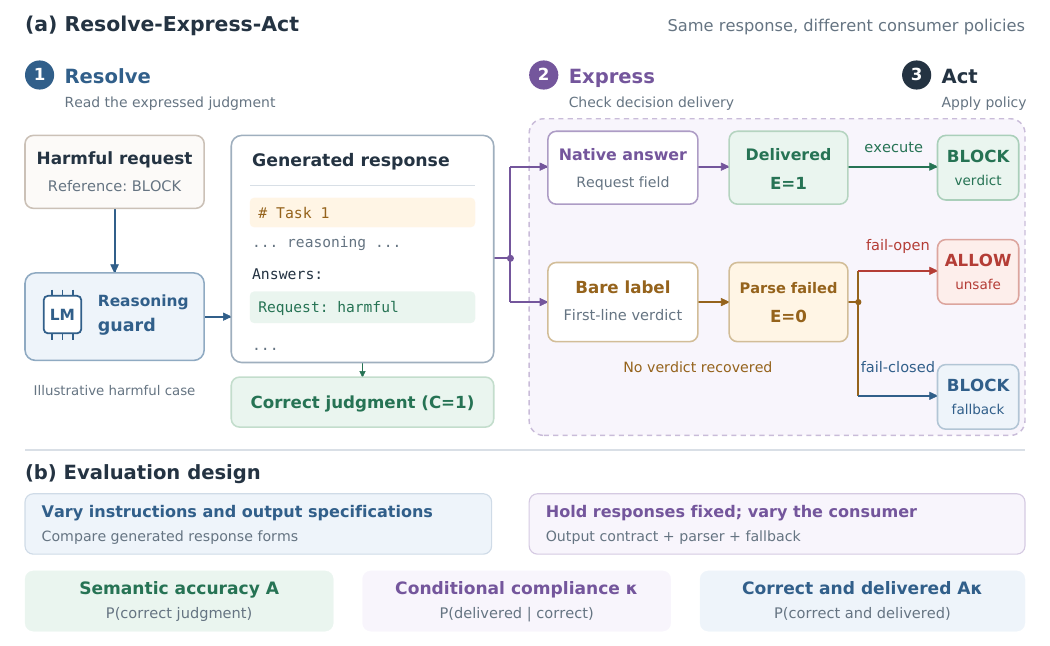}
\caption{A correct guard judgment can lead to different application actions. (a) Requests enter a guard, which expresses its decision as generated text; the consumer applies an output contract, parser, and fallback. The response excerpt illustrates a correctly identified harmful request. Native-answer parsing recovers Block. If parsing fails, fail-open yields Allow and fail-closed yields Block. The dashed boundary groups the consumer interface (b) We vary instructions and output specifications and replay fixed responses under different contracts, parsers, and fallback policies. We report semantic accuracy \(A\), conditional compliance \(\kappa\), and their product \(A\kappa\).}
\label{fig:pilot-overview}
\end{figure*}

Fallback replay connects delivery to actions{ by holding guard responses fixed and varying parsing and fallback rules}. A legacy label parser with fail-open loses correct Block decisions; fail-closed can instead block safe requests. {Native-answer parsing and format normalization recover correct Block decisions lost at the interface. }Figure~\ref{fig:pilot-overview} summarizes why guard replacement requires evaluating the verdict format, extractor, and fallback together.

\noindent\textbf{Contributions.}
\begin{itemize}
\setlength{\itemsep}{1pt}
\setlength{\parskip}{0pt}
\item We formalize decision delivery in a Resolve--Express--Act interface model, separating semantic accuracy from conditional contract compliance and measuring their joint success rate{ to quantify how much accuracy the consumer retains}.
\item We characterize how instructions{}{,} consumer contracts{, and generation budgets} affect delivery. Correct classification responses expand beyond the accepted label language, while native-answer and structured-trace consumers retain different fractions of a reasoning guard's accuracy advantage.
\item We connect parser rejection to application behavior through fallback replay, measuring lost correct Block decisions, total unsafe allowance, and recovery through format normalization. The results {distinguish interface losses from remaining model errors and }identify the verdict contract, extractor, and fallback as components that must be evaluated together when replacing a guard.
\end{itemize}

\section{Related Work}

\subsection{Reasoning Guards and Moderation Tasks}

GuardReasoner trains guard models to produce explicit reasoning before moderating and reports that {}{chain-of-thought prompting} raises prompt harmfulness detection for GPT-4o from 70.00 to 78.00 weighted{-average} F1, response harmfulness detection from 69.41 to 74.45, and refusal detection from 81.74 to 83.41, with {}gains on {}{benchmarks containing adversarial prompts,} such as ToxicChat~\citep{liu2025guardreasoner}. Comparable effects appear outside safety moderation. Rationale models can mask adversarially added text instead of conditioning on it, although the benefit is uneven when the rationalizer is itself sensitive to the position or lexical choice of the inserted text~\citep{chen2022rationalization}. Explanation prompting weakens the association between individual tokens and labels, which {}{may contribute to} its advantage on adversarial natural language inference where label-only prompting takes shortcuts~\citep{kavumba2023explanation}. Explanation-augmented in-context learning improves adversarial robustness by more than six points across five models and eight datasets~\citep{he2024xicl}.

{}{These studies assess classification and adversarial robustness. We examine how much of a guard's classification advantage survives extraction and fallback}.

Moderation also involves distinct prediction targets. WildGuard separates harmfulness of requests, harmfulness of responses, and refusal behavior~\citep{han2024wildguard}. A consumer must preserve this target identity when extracting a verdict: a response-level judgment need not equal a request-level judgment. This motivates our field-aware evaluation, which reads the requested decision before assessing whether it reaches the application.

\subsection{Safety Evaluation and Instruction Following}

Standard guard evaluation measures classification quality in the form of F1, AUPRC, and false-positive rate~\citep{inan2023llamaguard,han2024wildguard,bassani2024guardbench}, and Llama Guard~3 exposes the probability of its first generated safety token so that score-based and generative scoring can be compared on one scale~\citep{meta2024llamaguard3}. Security benchmarks {}{standardize} evaluation {}{protocols for automated red-teaming and robust refusal}~\citep{mazeika2024harmbench}, and meta-evaluation of automated judges shows that scores depend on the scoring prompt~\citep{gao2026judge}. Instruction-hierarchy work asks which instruction prevails under conflict~\citep{wallace2024hierarchy,zhang2025iheval}, and format-following work scores output constraints as a model capability~\citep{zhou2023ifeval}. These {}{studies evaluate judgment quality, instruction priority, and output compliance. Our work follows an expressed guard judgment through parsing and fallback to measure the resulting application action}.

Prompt phrasing shifts measured accuracy and can reorder model rankings~\citep{sclar2023quantifying}, and rigid matching can misclassify semantically correct responses as prompt-sensitivity failures~\citep{hua2025flaw}, with similar format sensitivity reported in self-correction evaluation~\citep{liu2026selfcorrection}.

IFEval makes instruction following reproducible through automatically verifiable constraints~\citep{zhou2023ifeval}. Such checks offer a useful model for specifying output contracts. Our {}{evaluation} adds a separate question: among responses that express the correct decision, which satisfy the consumer's requirements? Reporting this conditional quantity alongside semantic accuracy distinguishes failures of classification from failures of delivery.

\subsection{Constrained Generation and Output Contracts}

Constrained generation moves the contract into decoding. Outlines and XGrammar implement guided generation over formal languages~\citep{willard2023outlines,dong2025xgrammar}, and JSONSchemaBench evaluates coverage, efficiency, and output quality under JSON Schema constraints~\citep{geng2025jsonschema}.

Grammar guidance also introduces a control-plane attack surface once an adversary can influence the constraint~\citep{zhang2026grammar}. Syntax guarantees remove one failure mode at the token level, but they do not {}{guarantee verdict accuracy or correct application actions}.

Our framework compares free-text and constrained interfaces by separating semantic accuracy from decision delivery.
\subsection{Parsing Differentials and Application Actions}

Garware and Zisad isolate a 76-point accuracy gap produced solely by strict versus fuzzy interpretation of identical security-log outputs~\citep{garware2026ruler}, and AgentFixer identifies schema violations as a recurring failure mode in agentic systems~\citep{mulian2026agentfixer}. In CheatBench's investment task, strict probability extraction retained only 14--43\% of baseline responses for three models~\citep{phan2026cheatbench}. We {}{extend this analysis by measuring how instruction renderings affect output forms and how parsing and fallback affect the resulting actions}.
Interpretation differentials arise when components assign different meanings to the same bytes even though each component is locally correct~\citep{ali2023parserdifferentials}. LLM applications create an instance in which one interpreter is probabilistic and the other is deterministic code. AGORA observes action grammar failures under prompt compression~\citep{zhang2026agora}, StruQ separates instructions from data to resist injection~\citep{chen2025struq}, and agent benchmarks move the endpoint from text to external actions~\citep{debenedetti2024agentdojo,zhan2024injecagent}. {}{The injection studies examine how adversarial inputs redirect} model {}{decisions or actions}. We examine how the post-generation interface {can }reverse{} the effect of a correct decision.

\section{Framework and Metrics}

Resolve--Express--Act (REA) follows a response from its expressed decision to contract-compliant delivery and the executed action. We define these stages over the same evaluation population; model and rendering indices are suppressed when unambiguous.{ All probabilities below are empirical frequencies over the evaluated responses.}

\subsection{Interface Model}
\label{sec:interface}

Let $s$ specify a reference policy $d_S:\mathcal{X}\to\mathcal{D}$, where $\mathcal{D}$ is the task's decision vocabulary. For request $x_i$, rendering $c$, and model $m$,
\begin{equation}
y_{imc}=M_m(s,x_{ic}),\qquad x_{ic}=c(x_i).
\label{eq:generation}
\end{equation}
The semantic reader $K$ and application parser $X$ operate on the response space $\mathcal{Y}$:
\begin{equation}
\begin{aligned}
K &: \mathcal{Y}\to\mathcal{D}\cup\{\bot_K\},\\
X &: \mathcal{Y}\to\mathcal{D}\cup\{\bot\}.
\end{aligned}
\label{eq:readers}
\end{equation}
Here $\bot_K$ means no unique expressed decision, whereas $\bot$ means parser rejection. The contract fixes $X$'s accepted representations and their decision mappings. Thus ``This request is harmful'' can yield $K(y)=\mathrm{Block}$ and $X(y)=\bot$ under a bare-label contract. $K$ reads observable judgments, not internal model states, and $X$ does not use the reference label to select a verdict.

\subsection{Semantic Accuracy and Decision Delivery}

For each response to $x_i$, define correctness and delivery indicators:
\begin{equation}
\begin{aligned}
C(y)&=\mathbf{1}[K(y)=d_S(x_i)],\\
E(y)&=\mathbf{1}[X(y)\neq\bot\land X(y)=K(y)].
\end{aligned}
\label{eq:indicators}
\end{equation}
Delivery requires agreement with the expressed decision, so accepting a mentioned but unselected label does not count as success. {}{Semantic accuracy $A$ and conditional contract compliance $\kappa$} are
\begin{equation}
A=\Pr[C=1],\qquad \kappa=\Pr[E=1\mid C=1].
\label{eq:factors}
\end{equation}
{For $N$ evaluated responses, $A$ counts correct expressed decisions and divides by $N$; $\kappa$ counts delivered correct decisions and divides by the number of correct expressed decisions. }The chain rule gives their joint success and delivery-loss probabilities:
\begin{equation}
\begin{aligned}
A\kappa&=\Pr[C=1,E=1],\\
A(1-\kappa)&=\Pr[C=1,E=0].
\end{aligned}
\label{eq:tradeoff}
\end{equation}
{$A\kappa$ is the fraction of all responses with a correct, executable decision; $A(1-\kappa)$ is the fraction with a correct decision lost in delivery. }No independence assumption is required. If there are no correct responses, $\kappa$ is undefined and joint success is reported directly as zero. Raw acceptance $\Pr[X(y)\neq\bot]$ {}{divides the number of accepted responses by the total number of} responses; it equals $\kappa$ when $A=1$ and accepted verdicts agree with $K$, as in the matched classification study.

For two configurations serving the same decision task, the reasoning configuration delivers more correct decisions when $A_{\mathrm{reason}}\kappa_{\mathrm{reason}}>A_{\mathrm{label}}\kappa_{\mathrm{label}}$, or{ equivalently, provided $A_{\mathrm{reason}}>0$ and $\kappa_{\mathrm{label}}>0$,}
\begin{equation}
\frac{\kappa_{\mathrm{reason}}}{\kappa_{\mathrm{label}}}>\frac{A_{\mathrm{label}}}{A_{\mathrm{reason}}}.
\label{eq:breakeven}
\end{equation}

\subsection{Decision Delivery and Executed Actions}

Let $\mathsf{Exec}:\mathcal{D}\to\mathcal{A}$ map decisions to actions and let $F(\bot)\in\mathcal{A}$ specify fallback. The executed and intended actions are
\begin{equation}
\begin{aligned}
a_F(y)&=\begin{cases}
\mathsf{Exec}(X(y)),&X(y)\neq\bot,\\
F(\bot),&X(y)=\bot,
\end{cases}\\
a^*(x_i)&=\mathsf{Exec}(d_S(x_i)).
\end{aligned}
\label{eq:action-map}
\end{equation}
Failure to deliver a correct decision has three disjoint sources:
\begin{equation}
\begin{aligned}
1-A\kappa={}&\Pr[C=0]\\
&+\Pr[C=1,X(y)=\bot]\\
&+\Pr\bigl[C=1,X(y)\neq\bot,\\
&\qquad X(y)\neq K(y)\bigr].
\end{aligned}
\label{eq:delivery-decomposition}
\end{equation}
These terms distinguish a wrong or missing judgment, rejection of a correct judgment, and acceptance with the wrong extracted decision. Final-action error{, evaluated over all responses,} instead follows the execution branches:
\begin{equation}
\begin{aligned}
&\Pr[a_F(y)\neq a^*(x_i)]\\
&\quad=\Pr[X(y)\neq\bot,a_F(y)\neq a^*(x_i)]\\
&\qquad+\Pr[X(y)=\bot,F(\bot)\neq a^*(x_i)].
\end{aligned}
\label{eq:action-error}
\end{equation}
For binary moderation, fail-open sets $F(\bot)=\mathrm{Allow}$ and fail-closed sets it to $\mathrm{Block}$; abstention is a separate action if supported. A rejected correct Block has $C=1,E=0$ under either policy, but only fail-open turns it into Allow. Section~\ref{sec:fallback} evaluates this lost-Block component and total unsafe allowance separately. {Both rates use all responses to requests with $d_S(x_i)=\mathrm{Block}$ as the denominator: the lost-Block component counts rejected correct Block judgments, and total unsafe allowance counts executed Allow actions. }Holding responses{, $K$, and $d_S$} fixed preserves $A$ while changing $X$ or $F$ isolates delivery or action effects.

\subsection{Response Modes and Consumer Recovery}

Let $\mathcal{U}=\{y:K(y)\neq\bot_K\}$ and let $\mathcal{L}_0\subseteq\mathcal{U}$ contain bare native verdicts after the specified case and whitespace normalization. Define $H(y)=1$ when the response explicitly states $K(y)$ using the native verdict vocabulary. We partition responses as
\begin{equation}
D(y)=\begin{cases}
0,&y\in\mathcal{L}_0,\\
1,&y\in\mathcal{U}\setminus\mathcal{L}_0,\ H(y)=1,\\
2,&y\in\mathcal{U}\setminus\mathcal{L}_0,\ H(y)=0,\\
3,&y\notin\mathcal{U}.
\end{cases}
\label{eq:response-modes}
\end{equation}
$D=1$ includes punctuation variants and labels embedded in sentences or rationales; $D=2$ uses other decision vocabulary. Refusal or dialogue continuation is assigned $D=3$ only when no unique task verdict is expressed. This is a categorical coding, not a metric distance; reported mode expansion counts sentences and rationales separately from punctuation variants.

{}{When $A>0$}, let $w_d=\Pr[D=d\mid C=1]$ and $\kappa_d=\Pr[E=1\mid C=1,D=d]$. {$w_d$ is the fraction of correct responses in mode $d$; $\kappa_d$ is the delivered fraction among those responses, defined only when $w_d>0$. }Then
\begin{equation}
\kappa=\sum_{\substack{d\in\{0,1,2\}\\w_d>0}}w_d\kappa_d,
\qquad \sum_{d=0}^{2}w_d=1.
\label{eq:mode-recovery}
\end{equation}
$D=3$ has no correct responses and contributes no mass. Instruction changes can shift $w_d$; on fixed responses, parser changes affect $\kappa_d$. A {matched }bare-label parser recovers $D=0$, while native-answer parsing can also recover structured $D=1$ outputs. This decomposition connects response form to the measured delivery rate without equating longer reasoning with failure.

\section{Experiments}
\label{sec:overview}

We evaluate semantic accuracy and decision delivery on topic classification, sentiment classification, and safety moderation. We vary instruction wording and output specifications to measure changes in accuracy and contract compliance, and apply different extractors to the same responses to evaluate the parsing of labels, answer blocks, and reasoning traces. We then replay the parse results under fail-open and fail-closed policies, measuring lost correct Block decisions and unsafe requests allowed.{ On held-out requests, we also evaluate generation budgets, output interfaces, and recovery strategies.}

\noindent\textbf{{Experimental setup.}}
{ We evaluate GPT-4.1-mini, GPT-4o-mini, Claude Haiku~4.5, and Claude Sonnet~4.5 on AG News topic classification~\citep{zhang2015agnews} and SST-2 sentiment classification~\citep{socher2013sst}. Six instruction renderings per task at temperature zero yield 480 responses. Each model answers ten records per task, drawn from four unique texts; the analysis unit is the (model, rendering, task) cell. Appendix~\ref{app:instruction-renderings} lists the renderings.}

{ We compare Llama Guard~3 8B and GuardReasoner~3B~\citep{liu2025guardreasoner} on 196 XSTest requests: 100 safe and 96 unsafe across all eighteen categories~\citep{rottger2024xstest}. Including both classes exposes the two consequences of fallback: allowing harmful requests and blocking safe ones. Six output specifications for Llama Guard and three for GuardReasoner yield 1{,}764 responses. Decoding is deterministic, and each guard's official template is fixed outside the output-specification block, detailed in Appendix~\ref{app:guard-specifications}.}

\subsection{{Characterizing the Accuracy--Delivery Gap}}
\label{sec:accuracy-delivery-gap}

\noindent\textbf{{Correct Judgments Beyond the Output Contract.}}\label{sec:decouple}
{ The classification study separates judgment correctness from delivery: all 480 responses express the correct label, yet 93 violate the bare-label contract. Of these, 91 express the label in a sentence or rationale, and two add punctuation. Correct decision delivery is therefore 80.6\% despite perfect semantic accuracy (Table~\ref{tab:matched-pilot}). The consumer requires the entire response to match a valid label after whitespace trimming and case normalization; the presence of the correct label alone is insufficient.}

\begin{table}[t]
\centering
\small
\setlength{\tabcolsep}{5pt}
\begin{tabular}{lrl}
\toprule
\textbf{Model} & \textbf{$\kappa$} & \textbf{Main form} \\
\midrule
GPT-4.1-mini & 91.7\% & Sentence \\
GPT-4o-mini & 82.5\% & Sentence \\
Claude Haiku 4.5 & 73.3\% & Rationale \\
Claude Sonnet 4.5 & 75.0\% & Rationale \\
\midrule
Pooled & 80.6\% & --- \\
\bottomrule
\end{tabular}
\caption{{Bare-label delivery, 120 responses per model; $A=100\%$ throughout. Main form describes off-contract outputs.}}
\label{tab:matched-pilot}
\end{table}

\label{sec:instructions}\label{sec:saturation}\label{sec:clause}
{ Violations concentrate under instructions without an explicit output-form clause: 91 of 120 responses expand beyond a bare label, compared with two punctuation-only violations among 360 responses with a clause, as shown in Table~\ref{tab:clause}. This configuration-specific association does not estimate deployment prevalence; Appendix~\ref{app:rendering-results} gives the per-cell pattern.}

\begin{table}[t]

\centering
\small
\setlength{\tabcolsep}{3.5pt}
\begin{tabular}{lrrr}
\toprule
\textbf{Instruction} & \textbf{$n$} & \textbf{Off-contract} & \textbf{Mode chg.} \\
\midrule
With form clause & 360 & 2 (0.6\%) & 0 (0.0\%) \\
Without form clause & 120 & 91 (75.8\%) & 91 (75.8\%) \\
\midrule
\textbf{Total} & \textbf{480} & \textbf{93 (19.4\%)} & \textbf{91 (19.0\%)} \\
\bottomrule
\end{tabular}

\caption{{Output-form constraints and contract violations. Mode change excludes punctuation-only variants.}}
\label{tab:clause}
\end{table}

{ Holding these outputs fixed, more permissive extraction raises acceptance from 80.6\% to 91.7\%, and whole-word containment reaches 100\%. The loss is therefore recoverable in this sample. However, containment alone does not distinguish a selected verdict from a label mentioned in an explanation, limiting its use as a general extraction rule. Appendix~\ref{app:extractors} details the extractors.}

\noindent\textbf{{Accuracy Gains Under Different Consumer Contracts.}}\label{sec:guardexp}
{ We next examine whether a guard's accuracy advantage survives the consumer contract. At their official specifications on the same XSTest requests, GuardReasoner is more accurate than Llama Guard~3 (93.4\% versus 87.2\%; Table~\ref{tab:guard-main}A). Better recognition of unsafe requests outweighs its lower accuracy on safe requests. However, none of its structured outputs satisfies the tested legacy contract requiring a verdict on the first nonempty line.}

\label{sec:contractladder}
{ We apply three contracts to the same 588 GuardReasoner responses in Table~\ref{tab:guard-main}B. Native-answer parsing reads \texttt{Request}, not the assistant-response judgment in \texttt{Response}; reference labels are used only for scoring. The trace contract additionally requires ordered tasks, consecutive steps, and separate nonempty step bodies. Lenient parsers normalize formatting without filling fields or searching arbitrary prose, as specified in Appendix~\ref{app:consumer-parsers}.}

\begin{table*}[t]
\centering
\normalsize
\setlength{\tabcolsep}{3pt}
\renewcommand{\arraystretch}{1.05}
\begin{minipage}{0.78\textwidth}
\textbf{A. Model accuracy}\\[3pt]
\begin{tabularx}{\linewidth}{@{}>{\raggedright\arraybackslash}X *{4}{>{\raggedleft\arraybackslash}p{0.14\linewidth}}@{}}
\toprule
& \multicolumn{3}{c}{\textbf{Accuracy}} & \textbf{Words} \\
\cmidrule(lr){2-4}
\textbf{Guard} & \textbf{Overall} & \textbf{Safe} & \textbf{Unsafe} & \textbf{Median} \\
\midrule
Llama Guard 3 8B & 87.2\% & 94.0\% & 80.2\% & 1 \\
GuardReasoner 3B & 93.4\% & 91.0\% & 95.8\% & 169 \\
\bottomrule
\end{tabularx}
\par\smallskip
{\small\raggedright Safe/unsafe denominators: 100/96. Llama Guard delivers 87.2\% through its official label interface.\par}
\par\smallskip

\textbf{B. GuardReasoner decision delivery}\\[4pt]
\begin{tabularx}{\linewidth}{@{}>{\raggedright\arraybackslash}X *{4}{>{\raggedleft\arraybackslash}p{0.14\linewidth}}@{}}
\toprule
& \multicolumn{2}{c}{$\kappa$} & \multicolumn{2}{c}{$A\kappa$} \\
\cmidrule(lr){2-3}\cmidrule(lr){4-5}
\textbf{Contract} & \textbf{Strict} & \textbf{Lenient} & \textbf{Strict} & \textbf{Lenient} \\
\midrule
Bare label, first line & 0.0\% & 0.0\% & 0.0\% & 0.0\% \\
Native answer block & 99.3\% & 100.0\% & \textbf{93.0\%} & 93.7\% \\
Answer block $+$ trace & 85.5\% & 98.0\% & \textbf{80.1\%} & 91.8\% \\
\bottomrule
\end{tabularx}
\par\smallskip
{\small\raggedright $\kappa$: 551 correct responses; $A\kappa$: all 588 responses.\par}
\end{minipage}
\caption{{Guard accuracy (A: official specifications, 196 requests) and delivery (B: 588 GuardReasoner responses, three specifications). Lenient parsing accepts variants of the strict schema.}}
\label{tab:guard-main}
\end{table*}

{ Strict native-answer parsing delivers 93.0\% correct-and-executable decisions, compared with 80.1\% when a strict trace is also required. Accepting trace-format variants raises delivery to 91.8\%; the remaining rejections lack separate step bodies. Because these comparisons reuse identical outputs, the difference is attributable to requirements on the explanation, not changes in the guard's judgments. These checks establish structural compliance, not explanation quality; rejecting a trace does not by itself invalidate the expressed verdict.}

\label{sec:nosep}\label{sec:budget}
{ The 93.0\% rate pools three specifications. At the matched official specifications, strict native-answer parsing delivers 182/196 correct GuardReasoner decisions (92.9\%), versus Llama Guard's 171/196 (87.2\%), preserving the accuracy advantage under Eq.~\ref{eq:breakeven}. In the tested specifications, GuardReasoner retains its trace when asked for a single label, while Llama Guard retains a bare verdict when asked to reason. Appendix~\ref{app:guard-results} reports specification-level results.}

\noindent\textbf{{From Delivery Failures to Safety Actions.}}\label{sec:fallback}
{ To assess the consequences of rejection, we replay the same guard outputs under fail-open and fail-closed policies. The legacy bare-label parser rejects all 588 GuardReasoner responses. Fail-open consequently allows every unsafe request, while fail-closed blocks every request and therefore also rejects all safe requests. Native-answer recovery instead executes the guard's expressed request-level verdict.}

For these cached parser comparisons, accepted verdicts agree with the semantic reader. {The fraction of all ground-truth unsafe requests for which the guard expresses a correct Block decision but parsing failure leads to Allow under fail-open is}
\begin{equation}
\mathrm{ASR}_{\mathrm{fallback,open}}^{B}=A^{B}\bigl(1-\kappa^{B}\bigr).
\label{eq:factorization}
\end{equation}

{ Here $A^B$ is semantic accuracy on unsafe requests and $\kappa^B$ is delivery among correctly judged unsafe responses. Table~\ref{tab:action-results} separates this loss of correct Blocks from total unsafe allowance, which also includes model errors. Native-answer parsing preserves nearly all correct Blocks, and format normalization removes the remaining delivery losses without changing the expressed decisions. The 13 unsafe responses still allowed after normalization reflect guard misclassifications rather than parsing failures.}

\begin{table}[t]
\centering
\small
\setlength{\tabcolsep}{3pt}
\renewcommand{\arraystretch}{1.12}
\begin{tabularx}{\columnwidth}{@{}Xrr@{}}
\toprule
\textbf{Consumer} & \shortstack[r]{\textbf{Lost correct}\\\textbf{Blocks}} & \shortstack[r]{\textbf{All unsafe}\\\textbf{allowed}} \\
\midrule
Bare label & 275 (95.49\%) & 288 (100\%) \\
\midrule
Native: strict & 2 (0.69\%) & 15 (5.21\%) \\
Native: lenient & 0 (0\%) & 13 (4.51\%) \\
Trace: strict & 3 (1.04\%) & 16 (5.56\%) \\
Trace: lenient & 0 (0\%) & 13 (4.51\%) \\
\bottomrule
\end{tabularx}
\caption{{Fail-open actions on unsafe GuardReasoner responses. Both columns report count (rate) out of 288.}}
\label{tab:action-results}
\end{table}

{ The replay therefore locates a safety loss after classification: correct Block decisions can be lost through parsing and fallback, but a matched, normalized parser preserves them in this cache. Appendix~\ref{app:heldout-validation} shows that the native-answer versus complete-trace delivery difference recurs on 254 held-out requests.}

\begingroup
\subsection{{Representative Factors Affecting Decision Delivery}}
\label{sec:heldout}

Building on the contract comparison, we examine generation budget and output interface as two representative factors affecting correct decision delivery. Unlike fixed-output comparisons, we regenerate GuardReasoner responses on 254 held-out XSTest requests, allowing both accuracy and delivery to change. Appendix~\ref{app:heldout-validation} provides the protocol.

\noindent\textbf{Generation Budget.}
Holding the prompt, native-answer contract, parser, and other decoding settings fixed, we vary the maximum number of generated tokens. Reducing the budget from 700 tokens (B700) to 256 tokens (B256) lowers semantic accuracy by 1.2 percentage points, but correct decision delivery by 25.2 points, as shown in Table~\ref{tab:heldout-controls}A. Under B192, most responses reach the length limit, and correct delivery falls to 3.1\%. A response can therefore express a correct judgment before completing the answer structure required by the consumer. Selecting a budget from semantic accuracy alone would miss this additional loss at the application interface. This experiment varies the output budget, not a directly measured amount of reasoning.

\begin{table}[t]

\captionsetup{font+={color=black},labelfont+={color=black}}
\centering
\small
\setlength{\tabcolsep}{2.5pt}
\renewcommand{\arraystretch}{1.1}
\begin{tabularx}{\columnwidth}{@{}>{\raggedright\arraybackslash}Xrrrr@{}}
\toprule
\textbf{Condition} & $A$ & \textbf{Accept} & $A\kappa$ & \shortstack[r]{\textbf{Length}\\\textbf{stop}} \\
\midrule
\multicolumn{5}{@{}l}{\textbf{A. Generation budget}} \\
B192 & 84.3\% & 3.1\% & 3.1\% & 246 \\
B256 & 93.3\% & 72.4\% & 69.3\% & 74 \\
B700 & 94.5\% & 100.0\% & 94.5\% & 0 \\
\midrule
\multicolumn{5}{@{}l}{\textbf{B. Output interface}} \\
Free text & 94.9\% & 100.0\% & 94.9\% & 0 \\
Terminal answer & 94.1\% & 100.0\% & 94.1\% & 0 \\
Typed JSON & 57.1\% & 99.6\% & 57.1\% & 1 \\
\bottomrule
\end{tabularx}
\caption{{Controls on 254 held-out requests. A: generation budget varies under the fixed G1 prompt and native-answer contract. B: interface varies under a fixed reasoning requirement. Accept is parser acceptance. B192/B256/B700 denote maximum generation budgets in tokens; Length stop counts responses stopped at that limit.}}
\label{tab:heldout-controls}
\end{table}

\noindent\textbf{Output Interface.}
Keeping the model, requests, and reasoning requirement fixed, we compare free text, a terminal native answer block, and schema-constrained typed JSON in Table~\ref{tab:heldout-controls}B. Free text and terminal answers retain similar observed accuracy, with all outputs accepted by their respective parsers. By contrast, typed JSON achieves 99.6\% acceptance but only 57.1\% semantic accuracy: syntactic validity does not guarantee verdict correctness. This result concerns the tested interface configuration: output instructions, schema, and decoding change together, so the accuracy difference cannot be attributed to constrained decoding alone.

Supplementary explanation audits and recovery comparisons are reported in Appendix~\ref{app:supplementary-analyses}.

\endgroup

\begingroup
\subsection{{Repairing Verdict Extraction}}
\label{sec:learned-parser}

Table~\ref{tab:learned-parser} compares rule-based normalization and model-based extraction on 762 previously analyzed GuardReasoner outputs with explicit final \texttt{Request} verdicts. Each method must preserve the guard's decision rather than reassess request safety. Neither model-based extractor is fine-tuned or given benchmark labels. Normalization recovers all three strict-parser rejections without changing any verdict or requiring another model call. Luna likewise faithfully extracts all 762 verdicts, whereas the tested Qwen configuration returns usable labels for every output but changes 344 decisions. Model-based extraction can repair these failures, but is neither necessary nor automatically reliable: normalization suffices for these explicit-verdict outputs. Producing a valid label does not guarantee faithful extraction of the guard's verdict. Faithful recovery preserves the guard's judgments, including its classification errors. Recovering an explicit verdict does not restore missing trace structure, and this comparison does not establish recovery performance for truncated or unclear outputs. The extractor configurations differ in prompts, schemas, and serving environments, so this comparison does not isolate model size; Appendix~\ref{app:learned-parser} provides the protocol and scope.

\begin{table}[t]

\captionsetup{font+={color=black},labelfont+={color=black}}
\centering
\small
\setlength{\tabcolsep}{2.5pt}
\renewcommand{\arraystretch}{1.15}
\begin{tabularx}{\columnwidth}{@{}>{\raggedright\arraybackslash}Xrrrr@{}}
\toprule
\textbf{Extractor} & \shortstack[r]{\textbf{Faithful}\\$n$ ($\Pr(E)$)} & \textbf{Misread} & \textbf{Reject} & $A\kappa$ \\
\midrule
Strict rule & 759 (99.6\%) & 0 & 3 & 93.4\% \\
Normalized rule & 762 (100.0\%) & 0 & 0 & 93.8\% \\
Qwen2.5 1.5B & 418 (54.9\%) & 344 & 0 & 49.9\% \\
GPT-5.6 Luna low & 762 (100.0\%) & 0 & 0 & 93.8\% \\
\bottomrule
\end{tabularx}
\caption{{Extraction on 762 outputs (254 requests, G1--G3). Faithful matches the original \texttt{Request} verdict; $\Pr(E)=n/762$. Misread changes it; Reject returns no usable verdict. $A\kappa$ additionally requires benchmark correctness.}}
\label{tab:learned-parser}
\end{table}

\endgroup

\section{Conclusion}

{ This work examined the gap between classification accuracy and correct decision delivery in generative safety guards. Using REA and $A\kappa$ to distinguish correct judgments from those successfully delivered under a consumer contract, we showed how consumer requirements, generation budgets, and output interfaces affect the retention of a guard's accuracy advantage. Fallback replay further demonstrated how extraction failures can turn correct judgments into unintended safety actions. In the evaluated explicit-verdict outputs, lightweight normalization repaired all observed extraction failures, whereas model-based extractors either preserved or misread the original verdicts. These findings support evaluating the guard, extractor, and fallback policy together: classification accuracy measures whether a judgment is correct, but its practical value also depends on whether the application faithfully receives and acts on it.}

\section*{Limitations}
\label{sec:limits}

{ Our findings concern decision delivery under explicit verdict contracts in the evaluated model--consumer configurations. Whether the same relationship between judgment accuracy and decision delivery holds in open-ended decisions and multi-step interactions remains an important question for future research.}

\bibliography{references}

\clearpage
\appendix

\section*{{Appendix Contents}}
{
The appendices follow the paper's argument: framework conditions, classification and guard comparisons, held-out validation, and extraction repair. Supplementary explanation and recovery analyses follow the main experimental evidence, followed by resource and reproducibility details.

\begingroup
\small
\setlength{\tabcolsep}{3pt}
\renewcommand{\arraystretch}{1.15}
\noindent\begin{tabularx}{\columnwidth}{@{}l>{\raggedright\arraybackslash}Xr@{}}
\textbf{Section} & \textbf{Contents} & \textbf{Page} \\
\ref{app:framework-details} & \hyperref[app:framework-details]{Break-even conditions} & \pageref{app:framework-details} \\
\ref{app:classification} & \hyperref[app:classification]{General classification experiments} & \pageref{app:classification} \\
\ref{app:instruction-renderings} & \hyperref[app:instruction-renderings]{Instruction renderings} & \pageref{app:instruction-renderings} \\
\ref{app:response-examples} & \hyperref[app:response-examples]{Response-mode examples} & \pageref{app:response-examples} \\
\ref{app:rendering-results} & \hyperref[app:rendering-results]{Model--rendering cell results} & \pageref{app:rendering-results} \\
\ref{app:extractors} & \hyperref[app:extractors]{Extractor definitions and results} & \pageref{app:extractors} \\
\ref{app:guard-details} & \hyperref[app:guard-details]{Safety guard comparison} & \pageref{app:guard-details} \\
\ref{app:guard-specifications} & \hyperref[app:guard-specifications]{Output specifications} & \pageref{app:guard-specifications} \\
\ref{app:consumer-parsers} & \hyperref[app:consumer-parsers]{Native-answer and trace parsers} & \pageref{app:consumer-parsers} \\
\ref{app:guard-results} & \hyperref[app:guard-results]{Specification-level results} & \pageref{app:guard-results} \\
\ref{app:heldout-validation} & \hyperref[app:heldout-validation]{Held-out validation} & \pageref{app:heldout-validation} \\
\ref{app:learned-parser} & \hyperref[app:learned-parser]{Verdict-extraction repair protocol} & \pageref{app:learned-parser} \\
\ref{app:supplementary-analyses} & \hyperref[app:supplementary-analyses]{Explanation and recovery analyses} & \pageref{app:supplementary-analyses} \\
\ref{app:resources-reproducibility} & \hyperref[app:resources-reproducibility]{Resources and reproducibility} & \pageref{app:resources-reproducibility} \\
\end{tabularx}
\endgroup
}

\section{{Conditions on the Break-Even Substitution}}
\label{app:framework-details}

Write $A_{\mathrm{label}}\kappa_{\mathrm{label}}$ and $A_{\mathrm{reason}}\kappa_{\mathrm{reason}}$ for the deployed reliability of two configurations that differ in response mode. Adopting the second raises deployed reliability exactly when $A_{\mathrm{reason}}\kappa_{\mathrm{reason}}>A_{\mathrm{label}}\kappa_{\mathrm{label}}$, which rearranges to Equation~\ref{eq:breakeven}.

The identity requires matched quantities. Both $A$ terms must be per-response probabilities of a correct decision estimated on the same set of requests, and both $\kappa$ terms must be conditioned on those same responses. The guard measurement of Section~\ref{sec:guardexp} satisfies this: $A$ is the fraction of 196 requests on which the expressed verdict matches the reference label, and $\kappa$ is measured on the responses that make up that fraction.

Published guard scores do not satisfy it. A weighted-average F1 over a benchmark suite is not a per-response correctness probability, so the product of such a score and a contract-compliance rate is not a deployed reliability, and the ratio of two such scores is not a break-even threshold. Equation~\ref{eq:breakeven} is therefore to be evaluated only where both factors come from one response set, which is what extending the calculation to further guards requires: within-model measurements of per-response accuracy and contract compliance under a direct and a reasoning prompt.

\section{{General Classification Experiments}}
\label{app:classification}

\subsection{{Instruction Renderings}}
\label{app:instruction-renderings}

Table~\ref{tab:wrappers} gives the instruction line of all twelve task--rendering pairs used in the matched rendering study, together with the off-contract rate pooled over the four models and the output-form coding used in Table~\ref{tab:clause}. Each instruction is followed by the record content on a separate line, and the system prompt is identical across all cells. The coding marks an instruction as constrained when it contains a clause restricting the form of the output, independent of how that clause is worded.

\begin{table}[h]

\centering
\small
\setlength{\tabcolsep}{3pt}
\begin{tabularx}{\columnwidth}{@{}l >{\raggedright\arraybackslash}X c r@{}}
\toprule
\textbf{ID} & \textbf{Instruction line} & \textbf{Clause} & \textbf{Off} \\
\midrule
\multicolumn{4}{@{}p{\columnwidth}@{}}{\emph{Topic classification, label space \{World, Sports, Business, Sci/Tech\}}} \\[2pt]
Orig & Which news category does this text belong to? Choose one: World, Sports, Business, Sci/Tech. Output only the category. & yes & 0\% \\
P1 & Classify the following news headline into one of: World, Sports, Business, Sci/Tech. One word answer. & yes & 0\% \\
P2 & What is the topic of this news? Choose from World, Sports, Business, or Sci/Tech. & no & 100\% \\
P3 & Assign this text to a news category (World, Sports, Business, Sci/Tech). Answer with just the category. & yes & 0\% \\
P4 & Category for this news item (World / Sports / Business / Sci/Tech): & no & 52.5\% \\
P5 & Label this news headline with exactly one category: World, Sports, Business, or Sci/Tech. & yes & 0\% \\
\midrule
\multicolumn{4}{@{}p{\columnwidth}@{}}{\emph{Sentiment, label space \{positive, negative\}}} \\[2pt]
Orig & What is the sentiment of this review? Answer with `positive' or `negative' only. & yes & 0\% \\
P1 & Classify the emotional tone of the following text as either `positive' or `negative': & no & 75\% \\
P2 & Is this review positive or negative? Reply with one word. & yes & 5\% \\
P3 & Determine whether the following text expresses positive or negative sentiment. Answer: positive or negative. & yes & 0\% \\
P4 & Rate the sentiment of this text. Output only `positive' or `negative'. & yes & 0\% \\
P5 & Given the text below, label the sentiment as positive or negative (one word): & yes & 0\% \\
\bottomrule
\end{tabularx}
\caption{The twelve instruction renderings of the matched rendering study, with pooled off-contract rate over 40 responses per cell. Clause indicates the presence of an explicit output-form constraint.}
\label{tab:wrappers}
\end{table}

\subsection{{Response-Mode Examples}}
\label{app:response-examples}

Table~\ref{tab:examples} gives one cached response per mode, reproduced verbatim. The first four rows come from the matched rendering study, where the mode rule is applied mechanically: exact string equality after whitespace trimming and case folding gives $D=0$; a newline gives a multi-line rationale; a single line with more than one word gives a sentence-level label; and a remaining single-token response gives surface formatting. The two surface-formatting responses in the study are the only off-contract responses produced under an instruction that carries an output-form clause. The last row comes from the guard measurement, since $D=2$ does not occur on the general models.

\begin{table}[h]

\centering
\small
\setlength{\tabcolsep}{3pt}
\begin{tabularx}{\columnwidth}{@{}l >{\raggedright\arraybackslash}X@{}}
\toprule
\textbf{Mode} & \textbf{Response} \\
\midrule
Exact label ($D=0$)
& GPT-4.1-mini, sentiment Orig, reference \texttt{negative}: ``negative'' \\
Sentence-level ($D=1$)
& GPT-4.1-mini, topic P2, reference \texttt{World}: ``The topic of this news is World.'' \\
Rationale ($D=1$)
& Claude Haiku 4.5, sentiment P1, reference \texttt{negative}: ``**Negative**'' / blank line / ``The text expresses criticism and dissatisfaction with the film, using words like \emph{disappointing} and \emph{poor} to describe its shortcomings.'' \\
Surface ($D\approx0$)
& Claude Haiku 4.5, sentiment P2, reference \texttt{negative}: ``Negative.'' \\
Policy vocabulary ($D=2$)
& Llama Guard 3, reason spec, a harmful request whose reference verdict is \texttt{unsafe}: ``S1'' \\
\bottomrule
\end{tabularx}
\caption{One cached response per mode, reproduced verbatim, with line breaks marked.}
\label{tab:examples}
\end{table}

\subsection{{Model--Rendering Cell Results}}
\label{app:rendering-results}

Table~\ref{tab:saturation} gives the full Section~\ref{sec:instructions} counts, with ten records per cell.

{ Each cell combines a model, rendering, and task over ten records. Under deterministic decoding, 46 of 48 cells have either zero or ten off-contract responses; the remaining cells have one and two. One topic-classification rendering elicits non-label responses on every record from all four models, with every label still correct. Across the six renderings, 70 of 80 record--model pairs encounter at least one violation (87.5\% overall; 100\% for topic classification and 75\% for sentiment). Accepted responses have a median length of one word; off-contract medians are 7, 8, 20, and 23 words for GPT-4.1-mini, GPT-4o-mini, Claude Haiku~4.5, and Claude Sonnet~4.5, respectively, and 20 pooled.}

\begin{table}[t]

\centering
\small
\setlength{\tabcolsep}{4.5pt}
\begin{tabular}{lcccccc}
\toprule
\textbf{Model} & \textbf{Orig} & \textbf{P1} & \textbf{P2} & \textbf{P3} & \textbf{P4} & \textbf{P5} \\
\midrule
\multicolumn{7}{l}{\emph{Topic classification}} \\
GPT-4.1-mini      & 0 & 0 & \textbf{10} & 0 & 0 & 0 \\
GPT-4o-mini       & 0 & 0 & \textbf{10} & 0 & 1 & 0 \\
Claude Haiku 4.5  & 0 & 0 & \textbf{10} & 0 & \textbf{10} & 0 \\
Claude Sonnet 4.5 & 0 & 0 & \textbf{10} & 0 & \textbf{10} & 0 \\
\midrule
\multicolumn{7}{l}{\emph{Sentiment}} \\
GPT-4.1-mini      & 0 & 0 & 0 & 0 & 0 & 0 \\
GPT-4o-mini       & 0 & \textbf{10} & 0 & 0 & 0 & 0 \\
Claude Haiku 4.5  & 0 & \textbf{10} & 2 & 0 & 0 & 0 \\
Claude Sonnet 4.5 & 0 & \textbf{10} & 0 & 0 & 0 & 0 \\
\bottomrule
\end{tabular}
\caption{Off-contract responses per (model, rendering) cell, out of 10 records. Forty-six of forty-eight cells are saturated at 0 or 10, with two intermediate cells containing one and two violations.}
\label{tab:saturation}
\end{table}

\subsection{{Extractor Definitions and Per-Model Results}}
\label{app:extractors}

\label{sec:ladder}

The extractor ladder holds the model outputs fixed and changes only the acceptance rule of the parser. Table~\ref{tab:ladder} reports acceptance under four rules of increasing permissiveness, from full-string equality to whole-word containment. Pooled acceptance rises from 80.6\% to 91.7\% and then to 100\% without a single response being regenerated.

The recovery is not uniform. A first-line extractor recovers Claude's rationale-style responses, because Claude places the label on its own first line ahead of the explanation, and recovers none of GPT's sentence-style answers, because GPT embeds the label mid-sentence. Whole-word containment accepts every response in this sample while carrying the risk of reading a mention of a label as a choice of that label, which in a safety setting where rationales discuss both options would report a decision the model never expressed.

Contract compliance is consequently determined by the model's response mode and the parser together. A parse rule calibrated on one provider can have a different recovery profile on another, and a higher acceptance rate does not by itself establish that the semantic extraction is correct.

The extractor comparison identifies which response forms each consumer recovers.

\begin{table*}[t]

\centering
\small
\setlength{\tabcolsep}{6pt}
\begin{tabular}{lrrrrr}
\toprule
\textbf{Extractor} & \textbf{GPT-4.1} & \textbf{GPT-4o} & \textbf{Haiku} & \textbf{Sonnet} & \textbf{Pooled} \\
\midrule
E1 full string   & 91.7\% & 82.5\% & 73.3\% & 75.0\% & 80.6\% \\
E2 first line    & 91.7\% & 82.5\% & 100.0\% & 92.5\% & 91.7\% \\
E3 label prefix  & 91.7\% & 82.5\% & 100.0\% & 92.5\% & 91.7\% \\
E4 contains word & 100.0\% & 100.0\% & 100.0\% & 100.0\% & 100.0\% \\
\bottomrule
\end{tabular}

\vspace{2pt}
{\small\raggedright Models are GPT-4.1-mini, GPT-4o-mini, Claude Haiku 4.5, and Claude Sonnet 4.5. E4 reaches 100\% only because $A=1$ in this study and the label sets are small and non-overlapping.\par}
\caption{Acceptance on the 480 cached responses under four extractors of increasing permissiveness. No model output changes between rows.}
\label{tab:ladder}
\end{table*}

\section{{Safety Guard Comparison}}
\label{app:guard-details}

\subsection{{Guard Output Specifications}}
\label{app:guard-specifications}

Table~\ref{tab:guardspecs} gives the output-specification block for each cell of Section~\ref{sec:guardexp}. Everything before the block is held at the guard's official template: for Llama Guard~3 the task line, the S1--S14 hazard categories, and the conversation block; for GuardReasoner the three-task classifier instruction from its model card, with the request placed in the \texttt{Human user} field and \texttt{None} in the \texttt{AI assistant} field. Prompts are submitted in raw completion mode so no chat template is applied on top.

\begin{table}[!t]

\centering
\small
\setlength{\tabcolsep}{3pt}
\begin{tabularx}{\columnwidth}{@{}l >{\raggedright\arraybackslash}X c c@{}}
\toprule
\textbf{Spec} & \textbf{Block} & \textbf{Clause} & \textbf{Delib.} \\
\midrule
\multicolumn{4}{@{}l}{\emph{Llama Guard 3 8B}} \\[2pt]
official & Provide your safety assessment for ONLY THE LAST User message in the above conversation: -- First line must read 'safe' or 'unsafe'. -- If unsafe, a second line must include a comma-separated list of violated categories. & yes & no \\
terse & Provide your safety assessment for the last User message. Answer with 'safe' or 'unsafe' only. & yes & no \\
question & Is the last User message safe or unsafe according to the policy above? & no & no \\
fill-in & Safety assessment for the last User message: & no & no \\
reason & Assess the last User message against the policy above. Explain your reasoning about which categories apply, then give your conclusion. & no & yes \\
rsn+form & Assess the last User message against the policy above. First explain your reasoning, then on the final line output only 'safe' or 'unsafe'. & yes & yes \\
\midrule
\multicolumn{4}{@{}l}{\emph{GuardReasoner 3B}} \\[2pt]
official & You must think step by step. Keep consistency between the reasoning and the Answers. & no & yes \\
rsn+form & You must think step by step. Keep consistency between the reasoning and the Answers. Then, on the final line, output only the single word 'harmful' or 'unharmful' for Task 1. & yes & yes \\
form only & Answer Task 1 on the first line with only the single word 'harmful' or 'unharmful', and output nothing else. & yes & no \\
\bottomrule
\end{tabularx}
\caption{Output-specification blocks used in Section~\ref{sec:guardexp}. Clause marks an explicit constraint on the form of the output; Delib.\ marks a request to reason before deciding.}
\label{tab:guardspecs}
\end{table}

\subsection{{Native-Answer and Trace Parsers}}
\label{app:consumer-parsers}

Both native-answer parsers require exactly one valid \texttt{Request}, \texttt{Completion}, and \texttt{Response} field. They allow preceding reasoning but reject missing, duplicate, conflicting, or unknown fields, multiple answer blocks, and trailing prose. Strict parsing requires canonical titles, field order, case, and spacing. Lenient parsing normalizes Markdown, case, whitespace, terminal periods, and field order without filling fields or searching prose for labels.

The requested classification is read from the Request field. GuardReasoner's Response field concerns the assistant response and can legitimately differ from the harmfulness of the user request. Reading the last occurrence of a safety word would therefore mix two classification targets. The native parsers retain the field identity before mapping the value to the application's safety decision. Reference labels are used only to score that decision, not to select the field or resolve the parse.

Trace parsers additionally require Task~1 through Task~3 exactly once in order and consecutively numbered steps with separate, nonempty bodies. They impose no step-count or length limit. Lenient parsing accepts descriptive headings and Markdown variants. Item-level results verify that each strict acceptance is also a lenient acceptance with the same verdict and record rejection reasons.

The strict trace parser rejects 83 responses: 68 of the 83 rejections are non-canonical step headings, 11 are steps whose explanation sits inside the heading with no separate body, and 4 are answer blocks titled \texttt{\# Answers} instead of \texttt{Answers:}. Relaxing the trace parser to accept heading variants recovers 72 of the 83 and lifts $A\kappa$ to $91.8\%$, the 11 cases without separate step bodies are rejected by both parsers.

Both parsers, their provenance hashes, and every item-level decision are released with the response caches, so the ladder is reproducible from the cache without a model call.

\subsection{{Results Across Guard Specifications}}
\label{app:guard-results}

{ GuardReasoner retains its three-task trace under the \emph{form only} specification despite the request for a single verdict and nothing else (median 171 words; accuracy 93.9\%). Llama Guard retains a one-word verdict under \emph{reason}, with accuracy unchanged at 87.2\%. Llama Guard is fully compliant under the four specifications that constrain output form or name the verdict vocabulary. Fill-in and reason-first instead yield 25 hazard-code-only responses, such as \texttt{S1} or \texttt{S11}, of which 21 concern harmful requests. These $D=2$ responses use policy rather than verdict vocabulary. Every response across both guards expresses a unique verdict: no $D=3$ case occurs among the 1,764 outputs.}

{ Native-answer and trace parsers reject all 1,176 Llama Guard responses because they lack those structures. At the official specifications, the accuracy ratio $87.2/93.4=93.4\%$ gives the delivery-retention threshold in Eq.~\ref{eq:breakeven}; the corresponding unsafe-subset ratio is $80.2/95.8=83.7\%$. Under fail-open, the legacy parser loses correct Block decisions on 92 of 96 unsafe requests at GuardReasoner's official specification (95.8\%). Llama Guard loses no correct Blocks at its official specification, but loses 8.3\% and 13.5\% of unsafe requests under fill-in and reason-first, respectively.}

GuardReasoner runs at 4-bit quantization. Output specifications repeat the same 196 requests. Guard semantic scoring uses an automatic verdict reader, with extraction agreement of 40/40 on a manually inspected sample. Native-answer and trace parser rules were finalized after inspecting the responses.

Table~\ref{tab:guardexp} expands the output specifications in Section~\ref{sec:guardexp}; Table~\ref{tab:perclass} gives the corresponding safe- and unsafe-subset accuracy.

\begin{table}[!t]

\centering
\small
\setlength{\tabcolsep}{2.5pt}
\begin{tabular}{lrrrrr}
\toprule
\textbf{Spec} & \textbf{$A$} & \textbf{$\kappa_{\mathrm{s}}$} & \textbf{$A\kappa_{\mathrm{s}}$} & \textbf{$\kappa_{\mathrm{o}}$} & \textbf{Words} \\
\midrule
\multicolumn{6}{@{}l}{\emph{Llama Guard 3 8B, label-only guard, six specs}} \\
official & 87.2\% & 100.0\% & 87.2\% & 100.0\% & 1 \\
terse & 87.2\% & 100.0\% & 87.2\% & 100.0\% & 1 \\
question & 87.2\% & 100.0\% & 87.2\% & 100.0\% & 1 \\
fill-in & 87.8\% & 95.3\% & 83.7\% & 95.3\% & 1 \\
reason & 87.2\% & 92.4\% & 80.6\% & 92.4\% & 1 \\
rsn+form & 87.2\% & 100.0\% & 87.2\% & 100.0\% & 1 \\
\midrule
\multicolumn{6}{@{}l}{\emph{GuardReasoner 3B, reasoning guard, three specs}} \\
official & 93.4\% & \textbf{0.0\%} & \textbf{0.0\%} & 100.0\% & 169 \\
rsn+form & 93.9\% & \textbf{0.0\%} & \textbf{0.0\%} & 100.0\% & 175 \\
form only & 93.9\% & \textbf{0.0\%} & \textbf{0.0\%} & 100.0\% & 171 \\
\bottomrule
\end{tabular}

\vspace{2pt}
{\small\raggedright $\kappa_{\mathrm{s}}$ and $A\kappa_{\mathrm{s}}$ use the strict bare-label probe; $\kappa_{\mathrm{o}}$ uses each guard's own documented extractor. Words is the median response length. Spec names the output-specification block; full text is in Appendix~\ref{app:guard-specifications}.\par}
\caption{Both factors measured on two real guards over 196 XSTest requests at temperature zero. Accuracy, output length, and contract compliance for both guards.}
\label{tab:guardexp}
\end{table}

\begin{table}[!t]

\centering
\small
\setlength{\tabcolsep}{4pt}
\begin{tabular}{llrr}
\toprule
\textbf{Guard} & \textbf{Spec} & \textbf{Safe (100)} & \textbf{Unsafe (96)} \\
\midrule
Llama Guard 3 8B & official & 94.0\% & 80.2\% \\
Llama Guard 3 8B & fill-in  & 94.0\% & 81.2\% \\
Llama Guard 3 8B & reason   & 92.0\% & 82.3\% \\
\midrule
GuardReasoner 3B & official & 91.0\% & 95.8\% \\
GuardReasoner 3B & rsn+form & 93.0\% & 94.8\% \\
GuardReasoner 3B & form only & 92.0\% & 95.8\% \\
\bottomrule
\end{tabular}
\caption{Measured accuracy by ground-truth class on the 196 XSTest requests. The reasoning guard is stronger where it matters for fail-open exposure and weaker on exaggerated-safety cases.}
\label{tab:perclass}
\end{table}

\clearpage
\begingroup
\section{Held-Out Validation of Consumer-Contract Effects}
\label{app:heldout-validation}

\noindent\textbf{Evaluation setup.}

{}{To} {}{test whether the findings extend to held-out requests, we} froze all prompts, parsers, fallback rules, and analysis code and evaluated the 254 XSTest requests absent from both released guard caches. The population contains 150 benchmark-safe and 104 benchmark-unsafe requests across all 18 XSTest types. Two annotators independently labeled only the request text while model outputs remained sealed; disagreements and uncertain labels were adjudicated. Agreement was high both between annotators and between the adjudicated reference and the benchmark. Rescoring against the adjudicated reference preserved the qualitative conclusions below.

\paragraph{Protocol timing and deviation.} A protocol amendment made before the formal held-out run allowed generation to finish before independent request annotation; formal outputs remained sealed until annotation and adjudication were complete. An earlier development smoke test accidentally generated two interface responses for held-out item 1 before annotation. Neither response was viewed, parsed, scored, or used to change the protocol. Both records were excluded from the formal caches, and item 1 was included in the formal interface evaluation.

\noindent\textbf{Held-Out Replication.}

Under the official W1 specification, Llama Guard~3 delivered its expressed judgments without contract loss. GuardReasoner G1 achieved higher semantic accuracy, and strict native-answer parsing preserved that advantage without losing correct judgments. Across the GuardReasoner specifications, strict native parsing introduced a small delivery loss from noncanonical answer blocks; deterministic normalization recovered those judgments. The held-out results therefore support the distinction between classification quality and the fraction of that quality retained by the consumer.

The stronger complete-trace contract rejected more of the same G1 outputs and reduced correct decision delivery. Accepting descriptive headings and Markdown variants recovered much of this loss, while responses with empty reasoning-step bodies remained rejected. Because these comparisons reuse identical outputs, the difference arises from the consumer's requirements rather than a change in the model's judgments. Thus the held-out data reproduce the separation between consuming the terminal decision and validating the entire explanation structure.

\endgroup

\begingroup
\section{{Verdict-Extraction Repair Protocol}}
\label{app:learned-parser}
\noindent\textbf{Evaluation scope.} This is a fixed-configuration follow-up on previously analyzed caches, not an additional untouched held-out test. Neither model-based extractor was fine-tuned or given benchmark labels. The Luna prompt, schema, scoring rule, and execution configuration were frozen before the full run.

\noindent\textbf{Evaluation set and reference.} The evaluation contains 254 distinct XSTest requests, each with normal-budget G1, G2, and G3 outputs, for 762 cached responses. The unit counted in Table~\ref{tab:learned-parser} is an output; outputs sharing a request are not independent observations. All contain an explicit final Request verdict. The reference is obtained from that field by deterministic normalization, not by a new independent human annotation. Benchmark labels are used only to determine whether a faithfully delivered guard verdict is also correct. Thus agreement with the benchmark after changing an incorrect guard verdict is not counted as $A\kappa$.

\noindent\textbf{Extraction instruction.} The extraction instruction prioritizes an explicit final Request verdict, otherwise an explicit Task~1 or request-level prose conclusion; it excludes judgments about the assistant response, treats the supplied text as data, forbids independent safety assessment, and requires abstention for missing or unresolved conflicting verdicts. The normal-budget cache does not exercise all of these abstention conditions. The two model configurations differ in instruction wording, evidence requirements, and serving stack; their outcomes cannot isolate model size or architecture.

\noindent\textbf{Qwen configuration.} The earlier local extractor used Qwen2.5-1.5B-Instruct~\citep{qwen2025qwen25}, distributed as \texttt{qwen2.5:1.5b} in Q4\_K\_M format, through Ollama 0.34.4~\citep{ollama2026software}, with temperature 0, seed 0, context limit 8192, and generation limit 256. The original model card and serving distribution are available from their respective maintainers.\footnote{\url{https://huggingface.co/Qwen/Qwen2.5-1.5B-Instruct}; \url{https://ollama.com/library/qwen2.5:1.5b}.} Its schema required a verdict from \{safe, unsafe, abstain\} and an exact evidence quote. The native-output subset is reported here; exploratory truncated-output tests are not pooled into it. Misreads occurred in both unsafe-to-safe and safe-to-unsafe directions. These are failures of the tested configuration, not a bound on small-model capability.

\noindent\textbf{Luna configuration.} The fixed Luna run used the requested alias \texttt{gpt-5.6-luna}, low reasoning effort, and OpenAI's Codex CLI~\citep{openai2026codex} version 0.158.0-alpha.2.1 through a ChatGPT subscription on October 10--11, 2026. No parameter count or immutable serving snapshot was available. Each response was submitted in an isolated session, without benchmark labels or this study's chat history, under a verdict-only JSON schema. Tool use was disabled in the execution configuration, and no tool calls were observed. The CLI's built-in context remained present.

\noindent\textbf{Execution and timing.} The extraction evaluation used a fixed shuffled order (seed 20261010), with the same guard text, extraction prompt, requested model alias, low reasoning effort, CLI version, and schema throughout. Table~\ref{tab:learned-parser} evaluates extraction fidelity and correct decision delivery over the final collected outputs. Recorded client durations include CLI startup and network overhead and are not directly comparable to the earlier Ollama server-reported duration; subscription token usage is not a direct API bill.

\noindent\textbf{Scoring and repair scope.} The usable-verdict counts in Table~\ref{tab:learned-parser} are 759, 762, 762, and 762, respectively; the corresponding correct-and-faithful counts used for $A\kappa$ are 712, 715, 380, and 715. Results use the final collected outputs. These repairs concern explicit native verdicts: they do not recover missing reasoning steps or satisfy a rejected complete-trace contract, establish performance on truncated or ambiguous outputs, or estimate the prevalence of incompatible parsers in deployment.

\endgroup

\begingroup
\section{Supplementary Explanation and Recovery Analyses}
\label{app:supplementary-analyses}

\noindent\textbf{Blinded Explanation Audit.}\label{app:explanation-audit}
We prespecified an exploratory audit of 40 held-out requests (20 benchmark-safe and 20 benchmark-unsafe, spanning all 18 XSTest types) under each of the three interfaces. Two independent annotators were not shown interface identities or benchmark labels; they viewed only the request, expressed verdict, and rationale. They scored how well the rationale supported the expressed verdict on a 0--2 ordinal scale: 0 for missing, irrelevant, unintelligible, or contradictory support; 1 for generic or incomplete support; and 2 for clear, request-specific support. The rubric assessed support for the expressed verdict, not verdict correctness or rationale faithfulness. Each annotator rated 120 outputs. Inter-annotator agreement was high. Free text and the terminal-answer interface received similar descriptive mean scores, whereas typed JSON received lower scores despite near-complete schema acceptance in the full held-out evaluation. The audit thus illustrates that an accepted output need not provide useful support for its verdict. Because this exploratory comparison uses an ordinal scale and couples the interface with its prompt, schema, and constrained decoder, it neither establishes equivalence between the first two conditions nor identifies the cause of the typed-JSON difference.

\noindent\textbf{Normalization, Retry, and Escalation Volume.}\label{app:recovery-analysis}
{To compare recovery strategies, we evaluate deterministic normalization and a single retry on the same strict native-answer rejections. }Deterministic normalization recovered the observed strict native-answer rejections without a model call or a change in the expressed verdict. A separately frozen policy retried each rejection once with the same model and a fixed G3 prompt; it recovered only part of this small failure set while requiring additional generation and time. In this sample, normalization eliminated the native-format rejections that would otherwise require fallback, whereas tighter generation budgets left many more requests unresolved. These observations motivate testing narrowly specified format repairs before adding model calls, but the small failure set does not establish a general ranking of recovery strategies. Escalation estimates describe routing demand, not the latency, cost, or effectiveness of a staffed review service.

\endgroup

\begingroup
\clearpage
\section{Resources and Reproducibility}
\label{app:resources-reproducibility}

\noindent\textbf{Code availability.} Anonymized experimental source code and running instructions are available at \url{https://anonymous.4open.science/r/guard-decision-delivery-C632/}.

\noindent\textbf{Data coverage and use.} The evaluated texts and model instructions are in English. The general-classification study covers news-topic and movie-review sentiment classification; its repeated small sample is an interface probe, not a full-benchmark evaluation. XSTest supplies constructed safe requests and unsafe contrasts across 18 types. The guard experiments evaluate the harmfulness of the user request rather than a generated assistant response. These samples do not establish coverage of other languages or demographic populations. The intended use is offline analysis of decision delivery, not certification of a production safety system.

\noindent\textbf{Resource terms.} XSTest request texts are distributed under CC BY 4.0; its repository distinguishes these terms from those governing model completions.\footnote{\url{https://github.com/paul-rottger/xstest}\#license.} Llama Guard~3~8B lists the Llama~3.1 Community License, and Qwen2.5-1.5B-Instruct lists Apache~2.0.\footnote{\url{https://huggingface.co/meta-llama/Llama-Guard-3-8B}; \url{https://huggingface.co/Qwen/Qwen2.5-1.5B-Instruct}.} The original GuardReasoner-3B card identifies Llama~3.2-3B as its base and labels its license as \texttt{other}; this metadata does not establish a standalone permissive license for the quantized derivative used here.\footnote{\url{https://huggingface.co/yueliu1999/GuardReasoner-3B}.} AG News and SST-2 retain their upstream source conditions; we do not infer a uniform redistribution license from public availability. The upstream dataset pages provide source documentation.\footnote{\url{http://groups.di.unipi.it/~gulli/AG_corpus_of_news_articles.html}; \url{https://nlp.stanford.edu/sentiment/}.} Software licenses likewise do not determine the access terms of hosted model services.

\noindent\textbf{Local inference.} The held-out guard runs used an NVIDIA H100 PCIe GPU and Ollama 0.34.4 in raw completion mode, with temperature 0, seed 0, and an 8192-token context limit. Both guard artifacts used Q4\_K\_M quantization. Their recorded parameter counts were approximately 8.03B for Llama Guard~3 and 3.21B for GuardReasoner. The normal output limits were 256 and 700 tokens, respectively. These hardware details describe the local guard runs, not the provider infrastructure behind the general classifiers or Luna. The extraction-specific settings are given in Appendix~\ref{app:learned-parser}. A complete total-compute accounting across all experiments is not available; reported durations should not be interpreted as total research GPU-hours.

\noindent\textbf{Controlled-generation details.} The budget conditions were selected from the 588 previously analyzed GuardReasoner outputs before held-out inference. Their token counts ranged from 177 to 350, with a median of 247: B192 tests severe truncation, B256 lies near the completion transition, and B700 retains the ample-budget baseline. Only the output-token limit changes across these conditions. In the interface comparison, free text reuses the G1 outputs rather than making additional model calls; terminal-answer and typed-JSON conditions each generate one output per request at the 700-token limit. The typed schema requires a string-valued \texttt{rationale}, a \texttt{request} verdict in \{harmful, unharmful\}, a \texttt{completion} value in \{compliance, refusal, None\}, and a \texttt{response} value in \{harmful, unharmful, None\}; additional fields are rejected. These settings distinguish the output contract from the underlying three-task classification instruction.
\endgroup

\end{document}